\newif\ifnips
\nipsfalse
\ifnips
\documentclass{article}

    \usepackage[nonatbib,final]{neurips_2019}
    
\else

\documentclass[a4paper,twocolumn,10pt]{extarticle}
\usepackage{a4wide}
\usepackage{times}
\usepackage[margin=0.65in]{geometry}
\fi

\usepackage[utf8]{inputenc}
\usepackage{amsmath,amssymb}
\usepackage{booktabs}
\usepackage{color}
\usepackage{framed}
\usepackage{graphicx}
\usepackage{placeins}
\usepackage{soul}
\usepackage{url}
\usepackage{cleveref}

\makeatother
\definecolor{darkgreen}{RGB}{0, 140, 0}
\definecolor{antiquefuchsia}{rgb}{0.57, 0.36, 0.51}
\definecolor{auburn}{rgb}{0.43, 0.21, 0.1}
\definecolor{correct}{rgb}{0.91, 0.24, 0.22}

\newcommand{\etal}{\emph{et al.}}

\makeatletter
\renewcommand{\paragraph}{%
  \@startsection{paragraph}{4}%
  {\z@}{0.4em}{-1em}%
  {\normalfont\normalsize\bfseries}%
}

\title{Fixing the train-test resolution discrepancy}
\date{}
\author{Hugo Touvron, Andrea Vedaldi, Matthijs Douze, Herv\'e J\'egou \\
~\\
\scalebox{1.}{Facebook AI Research}
}

\begin{document}
\maketitle
\begin{abstract}
Data-augmentation is key to the training of neural networks for image classification. 
This paper first shows that existing augmentations induce a significant discrepancy between the size of the objects seen by the classifier at train and test time: 
in fact, a lower train resolution improves the classification at test time!

We then propose a simple strategy to optimize the classifier performance, that employs different train and test resolutions. 
It relies on a computationally cheap fine-tuning of the network at the test resolution. 
This enables training strong classifiers using small training images, and therefore significantly reduce the training time.
For instance, we obtain 77.1\% top-1 accuracy on ImageNet with a ResNet-50 trained on 128$\times$128 images, and 79.8\% with one trained at 224$\times$224.

A ResNeXt-101 32x48d pre-trained with weak supervision on 940 million  224$\times$224 images and further optimized with our technique for test resolution 320$\times$320 achieves 86.4\% top-1 accuracy (top-5: 98.0\%). 
To the best of our  knowledge this is the highest ImageNet single-crop accuracy  to date\footnote{\textbf{Update:} Since the publication of this paper at Neurips, we have improved this state of the art by applying our method to EfficientNet. 
See our note~\cite{touvron2020FixEfficientNet} for results and details.}.
\end{abstract}

\section{Introduction}\label{sec:introduction}

Convolutional Neural Networks~\cite{lecun1989backpropagation} (CNNs) are used extensively in computer vision tasks such as image classification~\cite{Krizhevsky2012AlexNet}, object detection~\cite{ren2015faster}, inpainting~\cite{xie2012image}, style transfer~\cite{gatys2016image} and even image compression~\cite{rippel2017real}.  
In order to obtain the best possible performance from these models, the training and testing data distributions should match.
However, often data pre-processing procedures are different for training and testing.
For instance, in image recognition the current best training practice is to extract a rectangle with random coordinates from the image, which artificially increases the amount of training data.
This region, which we call the \emph{Region of Classification} (RoC), is then resized to obtain a crop of a fixed size (in pixels) that is fed to the CNN.
At test time, the RoC is instead set to a square covering the central part of the image, which results in the extraction of a so called  ``center crop''.
This reflects the bias of photographers who tend center important visual content.
Thus, while the crops extracted at training and test time have the same size, they arise from different RoCs, which skews the distribution of data seen by the CNN.

\begin{figure*}[t]
\centering \includegraphics[width=0.7\linewidth]{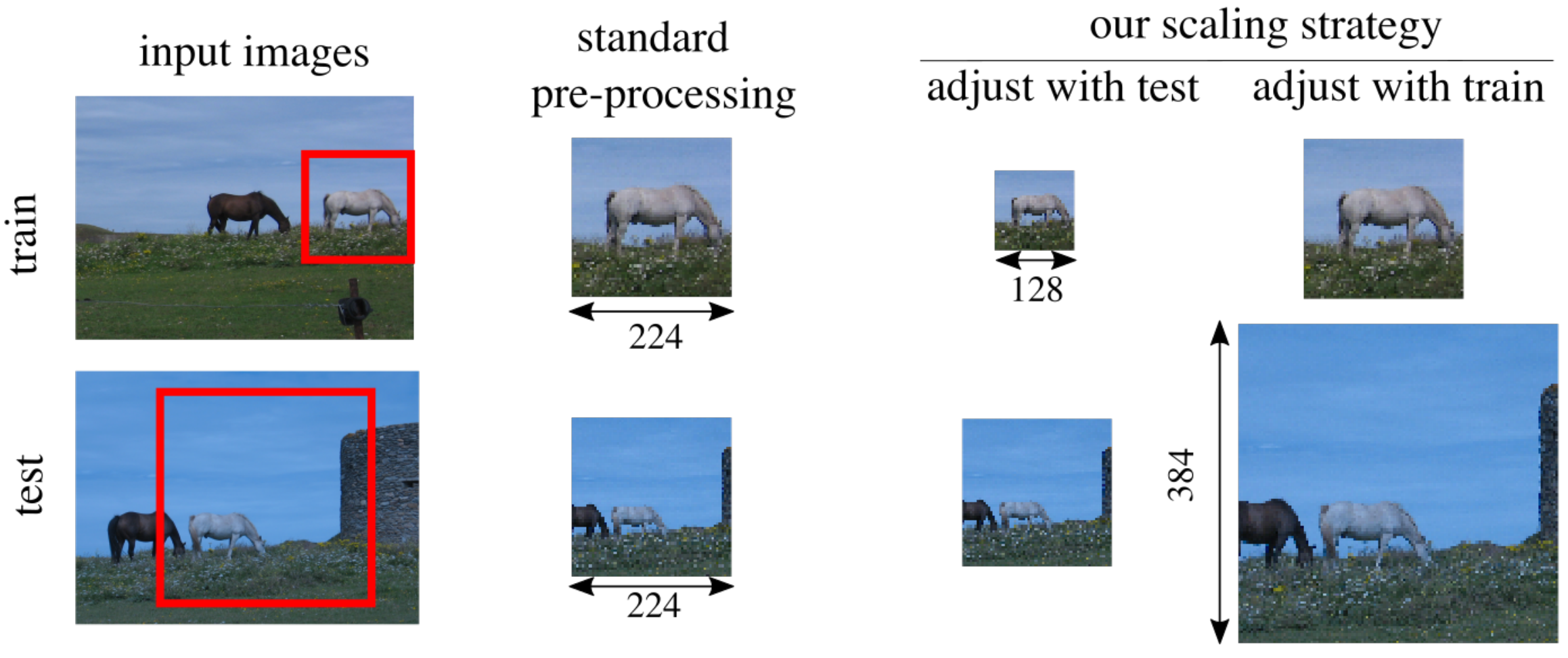}

\caption{\label{fig:traintestaug}
Selection of the image regions fed to the network at training time and testing time, with typical data-augmentation.
The red region of classification is resampled as a crop that is fed to the neural net.
For objects that have as similar size in the input image, like the white horse, the standard augmentations typically make them larger at training time than at test time (second column). 
To counter this effect, we either reduce the train-time resolution, or increase the test-time resolution (third and fourth column). The horse then has the same size at train and test time, requiring less scale invariance for the neural net. 
Our approach only needs a computationally cheap fine-tuning. }
\vspace{-0.2em}
\end{figure*}

Over the years, training and testing pre-processing procedures have evolved to improve the performance of CNNs, but so far they have been optimized separately~\cite{Ekin2018AutoAugment}.
In this paper, we first show that this separate optimization has led to a significant distribution shift between training and testing regimes with a detrimental effect on the test-time performance of models.
We then show that this problem can be solved by jointly optimizing the choice of resolutions and scales at training and test time, while keeping the same RoC sampling.
Our strategy only requires to fine-tune two layers in order to compensate for the shift in statistics caused by the changing the crop size.
This allows us to retain the advantages of existing pre-processing protocols for training and testing, including augmenting the training data, while compensating for the distribution shift.

Our approach is based on a rigorous analysis of the effect of pre-processing on the statistics of natural images, which shows that increasing the size of the crops used at test time compensates for randomly sampling the RoCs at training time. 
This analysis also shows that we need to use lower resolution crops at training than at test time. 
This significantly impacts the processing time: halving the crop resolution leads to a threefold reduction in the network evaluation speed and reduces significantly the memory consumption for a typical CNN, which is especially important for training on GPUs.
For instance, for a target test resolution of 224$\times$224, training at resolution 160$\times$160 provides better results than the standard practice of training at resolution 224$\times$224, while being more efficient. 
In addition we can adapt a ResNet-50 train at resolution 224$\times$224 for the test resolution 320$\times$320 and thus obtain top-1 accuracy of 79.8\% (single-crop) on ImageNet. 

Alternatively, we leverage the improved efficiency to train high-accuracy models that operate at much higher resolution at test time while still training quickly.
For instance, we achieve an top-1 accuracy of 86.4\% (single-crop) on ImageNet with a ResNeXt-101 32x48d pre-trained in weakly-supervised fashion on 940 million public images.  
Finally, our method makes it possible to save GPU memory, which could in turn be exploited by optimization: employing larger batch sizes usually leads to a better final performance~\cite{Tong2018BagofTricks}.

\section{Related work}\label{sec:related}

\paragraph{Image classification} is a core problem in computer vision. 
It is used as a benchmark task by the community to measure progress.
Models pre-trained for image classification, usually  on the ImageNet database~\cite{deng2009imagenet}, transfer to a variety of other applications~\cite{oquab2014learning}. 
Furthermore, advances in image classification translate to improved results on many other tasks~\cite{gordo2017end,Kornblith2018DoBetter}.

Recent research in image classification has demonstrated improved performance by considering larger networks and higher resolution images~\cite{Yanping2018GPipe,mahajan2018exploring}.
For instance, the state of the art in the ImageNet ILSVRC 2012 benchmark is currently held by the ResNeXt-101 32x48d~\cite{mahajan2018exploring} architecture with 829M parameters using 224$\times$224 images  for training.
The state of the art for a model learned  from  scratch  is  currently  held  by  the EfficientNet-b7~\cite{tan2019efficientnet} with 66M parameters using 600$\times$600 images  for training.
In this paper, we focus on the ResNet-50 architecture~\cite{He2016ResNet} due to its good accuracy/cost tradeoff (25.6M parameters) and its popularity.
We also conduct some experiments using the PNASNet-5-Large~\cite{Liu2018PNAS} architecture  that exhibits good performance on ImageNet with a reasonable training time and number of parameters (86.1M) and with the ResNeXt-101 32x48d~\cite{mahajan2018exploring} weakly supervised because it was the network publicly available with the best performance on ImageNet.

\paragraph{Data augmentation} is routinely employed at training time to improve model generalization and reduce overfitting. 
Typical transformations~\cite{Hu2017SENet,berman2019multigrain,Szegedy2015Goingdeeperwithconvolutions} include: random-size crop, horizontal flip and color jitter. 
In our paper, we adopt the standard set of augmentations commonly used in image classification.
As a reference, we consider the default models in the PyTorch library.
The accuracy is also improved by combining multiple data augmentations at test time, although this means that several forward passes are required to classify one image.
For example, \cite{He2016ResNet,Krizhevsky2012AlexNet,Szegedy2015Goingdeeperwithconvolutions} used ten crops (one central, and one for each corner of the image and their mirrored versions).
Another performance-boosting strategy is to classify an image by feeding it at multiple resolutions~\cite{He2016ResNet,Simonyan2015VGG,Szegedy2015Goingdeeperwithconvolutions}, again averaging the predictions. 
More recently, multi-scale strategies such as the feature pyramid network~\cite{lin2017feature} have been proposed to directly integrate multiple resolutions in the network, both at train and test time, with significant gains in category-level detection.

\paragraph{Feature pooling.}

A recent approach~\cite{berman2019multigrain} employs $p$-pooling instead of average pooling to adapt the network to test resolutions significantly higher than the training resolution.
The authors show that this improves the network's performance, in accordance with the conclusions drawn by Boureau~\etal~\cite{Boureau2010Pooling}. 
Similar pooling techniques have been employed in image retrieval for a few years~\cite{radenovic2018fine,tolias2015particular}, where high-resolution images are required to achieve a competitive performance. 
These pooling strategies are combined~\cite{tolias2015particular} or replace~\cite{radenovic2018fine} the RMAC pooling method~\cite{tolias2015particular}, which aggregates a set of regions extracted at lower resolutions.

%
%

\section{Region selection and scale statistics}\label{sec:analysis}

\ifnips 
\begin{figure}[t]
\centering
\begin{minipage}{0.48\linewidth}
\includegraphics[width=0.98\linewidth]{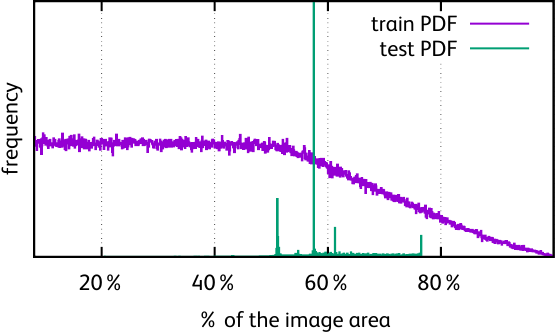}
\end{minipage}
\hfill
\raisebox{0.6em}
{
\begin{minipage}{0.48\linewidth}
\caption{\label{fig:pdfresolution}
Empirical distribution of the areas of the RoCs as a fraction of the image areas extracted by data augmentation. 
The data augmentation schemes are the standard ones used at training and testing time for CNN classifiers. 
The spiky distribution at test time is due to the fact that RoCs are center crops and the only remaining variability is due to the different image aspect ratios.
Notice that the distribution is very different at training and testing time.}
\end{minipage}}
\end{figure}

\else
\begin{figure}[t]
\includegraphics[width=0.95\linewidth]{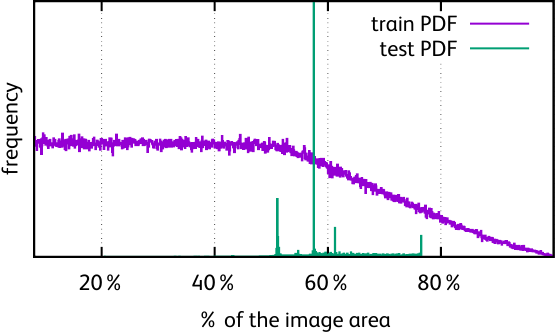}
\caption{\label{fig:pdfresolution}
Empirical distribution of the areas of the RoCs as a fraction of the image areas extracted by data augmentation. 
The data augmentation schemes are the standard ones used at training and testing time for CNN classifiers. 
The spiky distribution at test time is due to the fact that RoCs are center crops and the only remaining variability is due to the different image aspect ratios.
Notice that the distribution is very different at training and testing time.}
\end{figure}
\fi

Applying a Convolutional Neural Network (CNN) classifier to an image generally requires to pre-process the image.
One of the key steps involves selecting a rectangular region in the input image, which we call \emph{Region of Classification} (RoC).
The RoC is then extracted and resized to a square crop of a size compatible with the CNN, e.g.,~AlexNet requires a $224\times 224$ crop as input.

While this process is simple, in practice it has two subtle but significant effects on how the image data is presented to the CNN{}.
First, the resizing operation changes the \emph{apparent size} of the objects in the image (\cref{sec:apparent}).
This is important because CNNs \emph{do not} have a predictable response to a scale change (as opposed to translations).
Second, the choice of different crop sizes (for architectures such as ResNet that admit non-fixed inputs) has an effect on the \emph{statistics} of the network activations, especially after global pooling layers (\cref{sec:prelimactmap}).
This section analyses in detail these two effects.
In the discussion, we use the following conventions:
The ``input image'' is the original training or testing image;
the RoC is a rectangle in the input image;
and the ``crop'' is the pixels of the RoC, rescaled with bilinear interpolation to a fixed resolution, then fed to the CNN{}.

\subsection{Scale and apparent object size}\label{sec:apparent}

If a CNN is to acquire a scale-invariant behavior for object recognition, it must \emph{learn} it from data.
However, resizing the input images in pre-processing changes the  distribution of objects sizes.
Since different pre-processing protocols are used at training and testing time%
\footnote{
At training time, the extraction and resizing of the RoC is used as an opportunity to \emph{augment} the data by randomly altering the scale of the objects,
in this manner the CNN is stimulated to be invariant to a wider range of object scales.
},
the size distribution \emph{differs} in the two cases.
This is quantified next.

\subsubsection{Relation between apparent and actual object sizes}

We consider the following imaging model: the camera projects the 3D world onto a 2D image, so the apparent size of the objects is inversely proportional to their distance from the camera.
For simplicity, we model a 3D object as an upright square of height and width $R\times R$ at a distance $Z$ from the camera, and fronto-parallel to it.
Hence, its image is a $r \times r$ rectangle, where the \emph{apparent size} $r$ is given by $r=fR/Z$ where $f$ is the \emph{focal length} of the camera.
Thus we can express the apparent size as the product $r = f \cdot r_1$ of the focal length $f$, which depends on the camera, and of the variable $r_1=R/Z$, whose distribution $p(r_1)$ is camera-independent.
While the focal length is variable, the \emph{field of view} angle $\theta_{\text{FOV}}$ of most cameras is usually in the $[40^\circ, 60^\circ]$ range.
Hence, for an image of size $H\times W$ one can write  
$
 f = 
 k\sqrt{HW}
$
where
$
k^{-1} = 
2 \tan (\theta_{\text{FOV}}/2)
\approx 1
$
is approximately constant.
With this definition for $f$, the apparent size $r$ is expressed in pixels. 

\subsubsection{Effect of image pre-processing on the apparent object size}\label{s:aug}

Now, we consider the effect of rescaling images on the apparent size of objects.
If an object has an extent of $r\times r$ pixels in the input image, and if $s$ is the scaling factor between input image and the crop, then by the time the object is analysed by the CNN, it will have the new size of $rs \times rs$ pixels.
The scaling factor $s$ is determined by the pre-processing protocol, discussed next.

\paragraph{Train-time scale augmentation.}

\newcommand{\HWptr}{K_\text{train}}
\newcommand{\HWcte}{K_\text{test}^\text{image}}
\newcommand{\HWpte}{K_\text{test}}

As a prototypical augmentation protocol, we consider \texttt{RandomResizedCrop} in {PyTorch}, which is very similar to  augmentations used by other toolkits such as {Caffe} and the original AlexNet.
\texttt{RandomResizedCrop} takes as input an $H \times W$ image, selects a RoC at random, and resizes the latter to output a $\HWptr\times\HWptr$ crop.
The RoC extent is obtained by first sampling a scale parameter $\sigma$ such that $\sigma^2 \sim U([\sigma^2_-,\sigma^2_+])$ and an aspect ratio $\alpha$ such that $\ln \alpha \sim U([\ln\alpha_-,\ln\alpha_+])$.
Then, the size of the RoC in the input image is set to
$H_\text{RoC}\times W_\text{RoC} = \sqrt{\sigma^2\alpha HW} \times \sqrt{\sigma^2 HW/\alpha}$.
The RoC is resized anisotropically with factors $(\HWptr/H_\text{RoC}, \HWptr/W_\text{RoC})$ to generate the output image.
Assuming for simplicity that the input image is square (i.e.~$H=W$) and that $\alpha=1$, the scaling factor from input image to output crop is given by:
\begin{equation}\label{e:randmresized}
		s
		=
		\frac
		{\sqrt{\HWptr \HWptr}}{\sqrt{H_\text{RoC} W_\text{RoC}}}
		%
		%
		%
		%
	  =		
		\frac{1}{\sigma}
		\cdot
		\frac{\HWptr}{\sqrt{HW}}.
\end{equation}
By scaling the image in this manner, the apparent size of the object becomes
\begin{equation}\label{e:raug}
 r_\text{train}
 = s \cdot r
 = s f \cdot r_1 
 =  
 \frac{k\HWptr}{\sigma}
 \cdot
 r_1.
\end{equation}
Since $k\HWptr$ is constant, differently from $r$, $r_\text{train}$ does \emph{not} depend on the size $H\times W$ of the input image.
Hence, pre-processing \emph{standardizes} the apparent size, which otherwise would depend on the input image resolution.
This is important as networks do not have built-in scale invariance.

\paragraph{Test-time scale augmentation.}

As noted above, test-time augmentation usually differs from train-time augmentation.
The former usually amounts to: isotropically resizing the image so that the shorter dimension is $\HWcte$ and then extracting a $\HWpte \times \HWpte$ crop (\texttt{CenterCrop}) from that.
Under the assumption that the input image is square ($H=W$), the scaling factor from input image to crop rewrites as $s = \HWcte/\sqrt{HW}$, so that
\begin{equation}\label{e:teaug}
r_{\text{test}} = 
s\cdot r =
k \HWcte \cdot r_1.
\end{equation}
This has a a similar size standardization effect as the train-time augmentation.

\paragraph{Lack of calibration.}

Comparing~\cref{e:raug,e:teaug}, we conclude that the same input image containing an object of size $r_1$ results in two different apparent sizes if training or testing pre-processing is used.
These two sizes are related by:
\begin{align}
	\frac{r_{\text{test}}}{r_{\text{train}}} =
	\sigma \cdot \frac{\HWcte}{\HWptr}.
	\label{eq:apparentsizeratio}
\end{align}
In practice, for standard networks such as AlexNet $\HWcte/\HWptr \approx 1.15$; however, the scaling factor $\sigma$ is sampled (with the square law seen above) in a range $[\sigma_-, \sigma_+] = [0.28, 1]$.
Hence, at testing time the same object may appear as small as a third of what it appears at training time.
For standard values of the pre-processing parameters, the expected value of this ratio w.r.t.~$\sigma$ is
\begin{equation}
	\operatorname{E}\left[\frac{r_{\text{test}}}{r_{\text{train}}}\right] =
    F\cdot 
		\frac{\HWcte}{\HWptr}
		\approx 0.80,
		\qquad
		F = \frac{2}{3}
    \cdot
    \frac{\sigma_+^3 - \sigma_-^3}
    {\sigma_+^2 - \sigma_-^2},
	\label{eq:avgscaleratio}
\end{equation}
where $F$ captures all the sampling parameters.

\subsection{Scale and activation statistics}\label{sec:prelimactmap}

In addition to affecting the apparent size of objects, pre-processing also affects the activation statistics of the CNN, especially if its architecture allows changing the size of the input crop. %
We first look at the \emph{receptive field size} of a CNN activation in the previous layer.
This is the number of input spatial locations that affect that response.
For the convolutional part of the CNN, comprising linear convolution, subsampling, ReLU, and similar layers, changing the input crop size is almost neutral because the receptive field is unaffected by the input size.
However, for classification the network must be terminated by a pooling operator (usually average pooling) in order to produce a fixed-size vector. %
Changing the size of the input crop strongly affects the activation statistics of this layer.

\paragraph{Activation statistics.}
We measure the distribution of activation values after the average pooling in a ResNet-50 in~\cref{fig:pdfbatchnormeffect}. 
As it is applied on a ReLU output, all values are non-negative.
At the default crop resolution of $\HWpte$\,=\,$\HWptr$\,=\,224 pixels, the activation map is 7$\times$7 with a depth of 2048.
At $\HWpte$\,=\,64, the activation map is only 2$\times$2: pooling only 0 values becomes more likely and  activations are more sparse (the rate of 0's increases form 0.5\% to 29.8\%).
The values are also more spread out: the fraction of values above 2 increases from 1.2\% to 11.9\%.
Increasing the resolution reverts the effect: with $\HWpte$\,=\,448, the activation map is 14$\times$14,  the output is less sparse and less spread out. 

This simple statistical observations shows that
if the distribution of activations changes at test time, the values are not in the range that the final classifier layers (linear \& softmax) were trained for.

\subsection{Larger test crops result in better accuracy}\label{sec:paradox}

Despite the fact that increasing the crop size affects the activation statistics, it is generally beneficial for accuracy, since as discussed before it reduces the train-test object size mismatch.
For instance, the accuracy of ResNet-50 on the ImageNet validation set as $\HWpte$ is changed (see~\cref{sec:experiments}) are:

{\footnotesize
\ifnips
    ~ \hfill \begin{tabular}{lrrrrrrrr}
    \toprule
        $\HWpte$ & 64 & 128 & 224 & 256 & 288 & 320 & 384 & 448 \\
        accuracy & 29.4 & 65.4 & 77.0 & 78.0 & 78.4 & 78.3 & 77.7 & 76.6 \\
    \bottomrule
    \end{tabular} \hfill ~

\else
    \begin{center}
    \begin{tabular}{lr@{\hspace{10pt}}r@{\hspace{10pt}}r@{\hspace{10pt}}r@{\hspace{10pt}}r@{\hspace{10pt}}r@{\hspace{10pt}}r@{\hspace{10pt}}r@{\hspace{10pt}}}
    \toprule
        $\HWpte$ & 64 & 128 & 224 & 256 & 288 & 320 & 384 & 448 \\
        accuracy & 29.4 & 65.4 & 77.0 & 78.0 & 78.4 & 78.3 & 77.7 & 76.6 \\
    \bottomrule
    \end{tabular} 
    \end{center}
\fi
}

Thus for $\HWpte=288$ the accuracy is 78.4\%, which is \emph{greater} than 77.0\% obtained for the native crop size $\HWpte=\HWptr=224$ used in training.
In \cref{fig:accuracyresolution}, we see this result is general: better accuracy is obtained with higher resolution crops at test time than at train time.
In the next section, we explain and leverage this discrepancy by adjusting the network's weights.

\ifnips
    \begin{figure*}
    \begin{minipage}{0.45\linewidth}
    \includegraphics[width=0.98\linewidth]{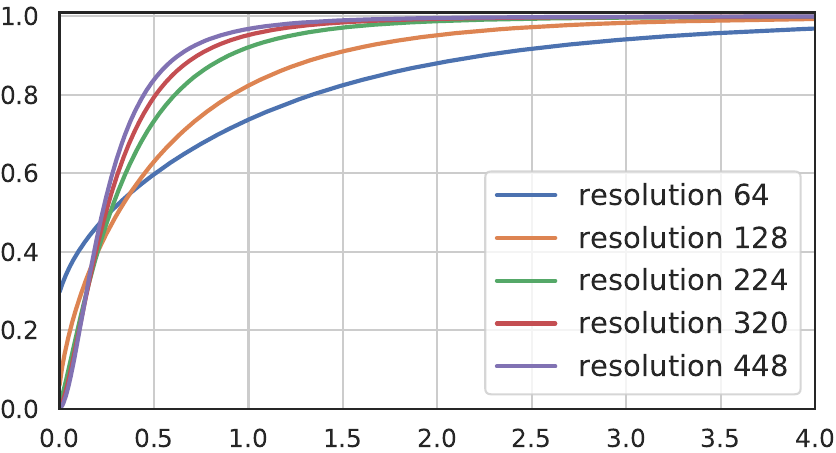}
    \end{minipage}
    \hfill
    \raisebox{1.8em}{
    \begin{minipage}{0.48\linewidth}
    \caption{\label{fig:pdfbatchnormeffect}
        Cumulative density function of the vectors components on output of the spatial average pooling operator, 
        for  a standard ResNet-50 trained at resolution  224, and tested at different resolutions. 
        The distribution is measured on the validation images of Imagenet. 
    }
    \end{minipage}}
    \vspace{-0.1em}
    \end{figure*}
\else
    \begin{figure}
    \includegraphics[width=0.98\linewidth]{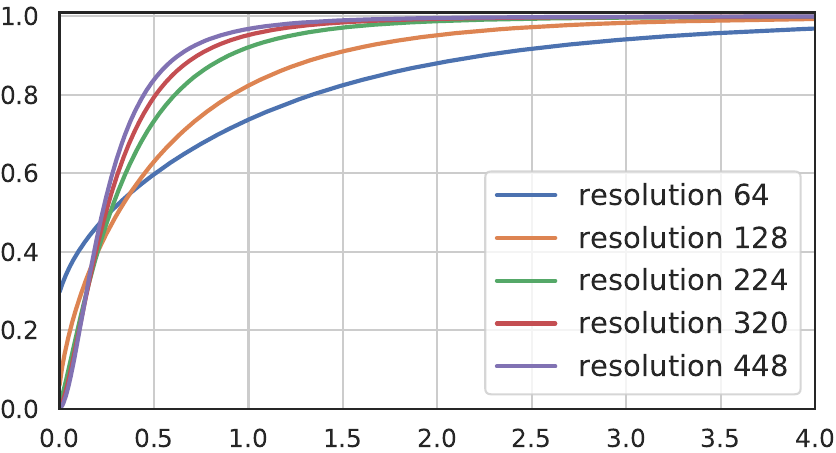}
    \caption{\label{fig:pdfbatchnormeffect}
        Cumulative density function of the vectors components on output of the spatial average pooling operator, 
        for  a standard ResNet-50 trained at resolution  224, and tested at different resolutions. 
        The distribution is measured on the validation images of Imagenet. 
    }
    \end{figure}
\fi

\section{Method}\label{sec:approach}

Based on the analysis of~\cref{sec:analysis}, we propose two improvements to the standard setting. 
First, we show that the difference in apparent object sizes at training and testing time can be removed by increasing the crop size at test time, which explains the empirical observation of~\cref{sec:paradox}.
Second, we slightly adjust the network before the global average pooling layer in order to compensate for the change in activation statistics due to the increased size of the input crop.

\subsection{Calibrating the object sizes by adjusting the crop size}

\Cref{eq:avgscaleratio} estimates the change in the apparent object sizes during training and testing.
If the size of the intermediate image $\HWcte$ is \emph{increased} by a factor $\alpha$ (where $\alpha \approx 1/0.80 = 1.25$ in the example) then
at test time, the apparent size of the objects is increased by the same factor. 
This equalizes the effect of the training pre-processing that tends to zoom on the objects.
However, increasing $\HWcte$ with $\HWpte$ fixed means looking at a smaller part of the object.
This is not ideal: the object to identify is often well framed by the photographer, so the crop may show only a detail of the object or miss it altogether.
Hence, in addition to increasing $\HWcte$, we \emph{also} increase the crop size $\HWpte$ to keep the ratio $\HWcte / \HWpte$ constant. %
However, this means that $\HWpte > \HWptr$, which skews %
the activation statistics (\cref{sec:prelimactmap}).
The next section shows how to compensate for this skew.

\ifnips
\subsection{Adjusting the statistics before spatial pooling}
\else
\subsection{Adjusting statistics before spatial pooling}
\fi

At this point, we have selected the ``correct'' test resolution for the crop but we have skewed activation statistics.
Hereafter we explore two approaches to compensate for this skew. %

\paragraph{Parametric adaptation.}

We fit the output of the average pooling layer (\cref{sec:prelimactmap}) with a parametric Fr\'echet distribution at the original $\HWptr$ and final $\HWpte$ resolutions.
Then, we define an equalization mapping from the new distribution back to the old one via a scalar transformation, and apply it as an activation function after the pooling layer (see Appendix~\ref{sec:parametricfit}).
This compensation provides a measurable but limited improvement on accuracy, probably because the model is too simple and does not differentiate the distributions of different components going through the pooling operator.

\paragraph{Adaptation via fine-tuning.} 

Increasing the crop resolution at test time is effectively a domain shift.
A natural way to compensate for this shift is to fine-tune the model.
In our case, we fine-tune on the same training set,  after switching from $\HWptr$ to $\HWpte$.
Here we choose to restrict the fine-tuning to the very last layers of the network.

A take-away from the distribution analysis is that the sparsity should be adapted. This requires at least to include the batch normalization that precedes the global pooling into the fine-tuning.
In this way the batch statistics are adapted to the increased resolution. 
We also use the test-time augmentation scheme during fine-tuning to avoid incurring further domain shifts.

\begin{figure*}
\begin{center}
\begin{minipage}{0.24\linewidth}
    \includegraphics[width=\columnwidth]{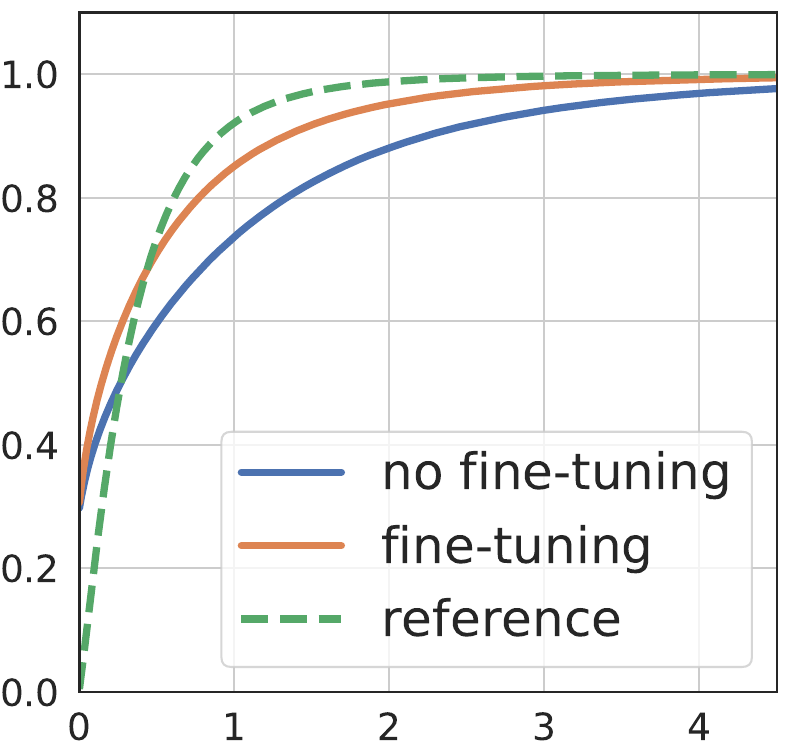} \\
    \centering $\HWpte=64$
\end{minipage}
\hfill
\begin{minipage}{0.24\linewidth}
    \includegraphics[width=\columnwidth]{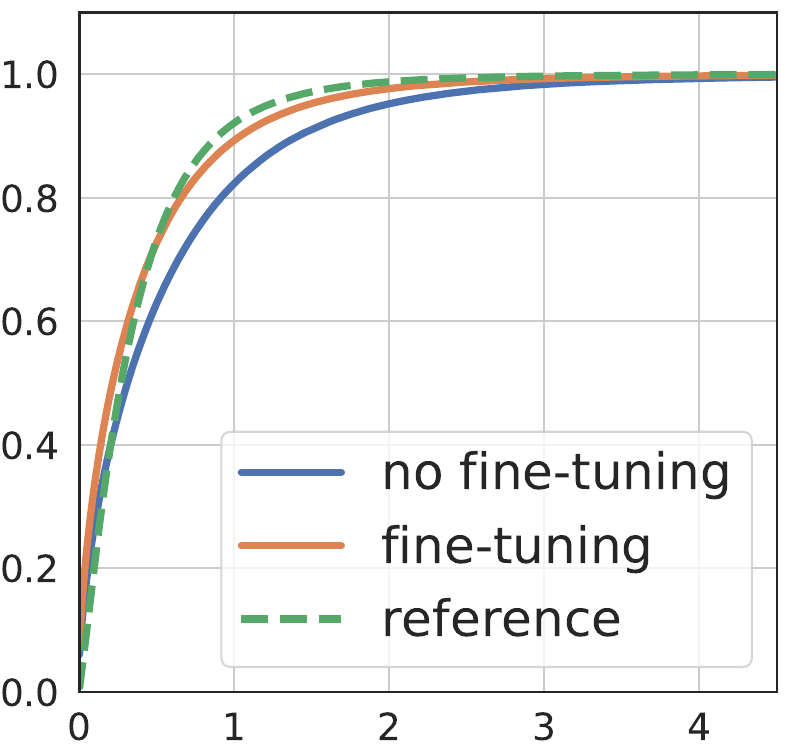} \\
    \centering $\HWpte=128$
\end{minipage}
\hfill
\begin{minipage}{0.24\linewidth}
    \includegraphics[width=\columnwidth]{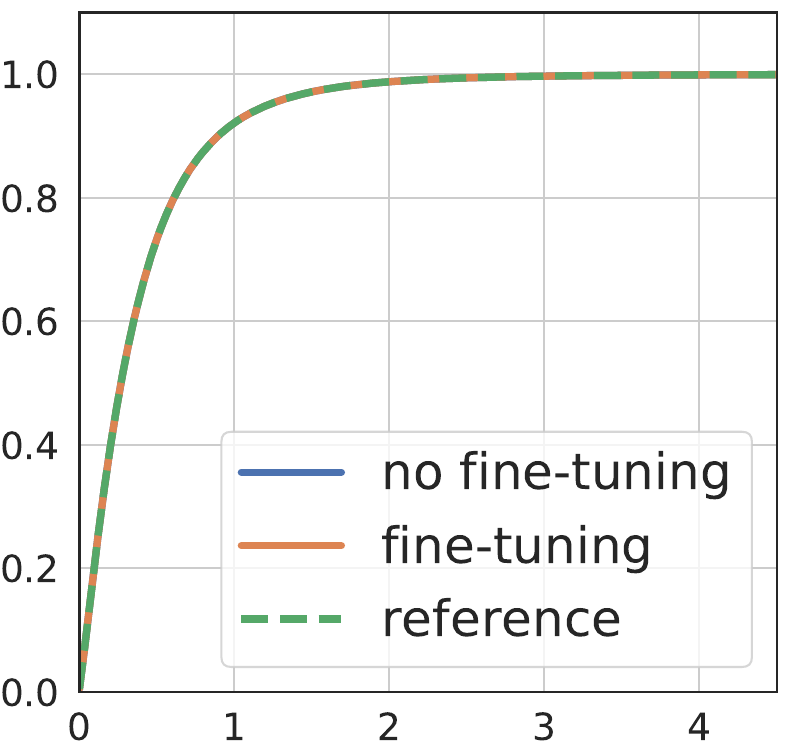} \\
    \centering $\HWpte=224$
  \end{minipage}
\hfill
\begin{minipage}{0.24\linewidth}
    \includegraphics[width=\columnwidth]{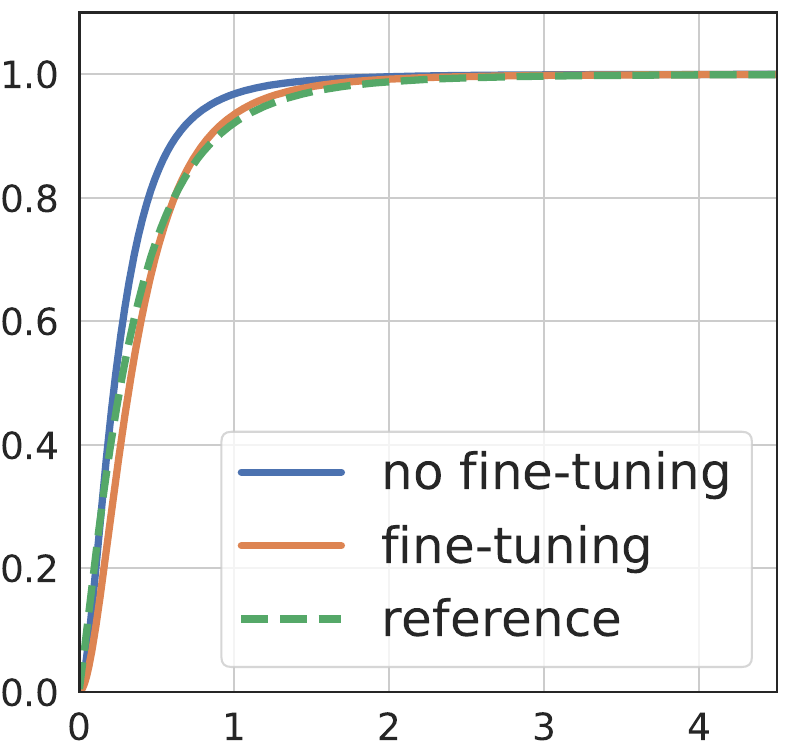}\\
    \centering $\HWpte=448$ 
\end{minipage}
\end{center}
\caption{\label{fig:cdfbatchnormeffect}
    CDF of the activations on output of the average pooling layer, for a ResNet-50, when tested at different resolutions $\HWpte$. 
    Compare the state before and after fine-tuning the batch-norm.
}
\end{figure*}

\Cref{fig:cdfbatchnormeffect} shows the pooling operator's activation statistics before and after fine-tuning.
After fine-tuning the activation statistics closely resemble the train-time statistics. %
This hints that adaptation is successful.
However, as discussed above, this does not imply an improvement in accuracy.

\section{Experiments}\label{sec:experiments}

\newcommand{\bs}{B}

\label{sec:setup}

\paragraph{Benchmark data.}

We experiment on the ImageNet-2012 benchmark~\cite{Russakovsky2015ImageNet12},
reporting validation performance as top-1 accuracy. 
It has been argued that this measure is sensitive to errors in the ImageNet labels~\cite{stock2018convnets}. 
However, the top-5 metrics, which is more robust, tends to saturate with modern architectures, while the top-1 accuracy is more sensitive to improvements in the model.

To assess the significance of our results, we compute the standard deviation of the top-1 accuracy: 
we classify the validation images,
split the set into 10 folds and measure the accuracy on 9 of them, leaving one out in turn. 
The standard deviation of accuracy over these folds is $\sim 0.03\%$ for all settings. 
\ifnips
    Therefore, we report 1 significant digit in the accuracy percentages. 
\else 
    Thus we report 1 significant digit in the accuracy percentages.
\fi

In the supplemental material, we also report results on the Fine-Grained Visual Categorization challenges iNaturalist and Herbarium. 

\paragraph{Architectures.}

We use standard state-of-the-art neural network architectures with no modifications, 
We consider in particular ResNet-50~\cite{He2016ResNet}. 
For larger experiments, we use PNASNet-5-Large \cite{Liu2018PNAS}, learned using ``neural architecture search'' as a succession of interconnected cells. 
It is accurate (82.9\% Top-1) with relatively few parameters (86.1 M).
We use also  ResNeXt-101 32x48d  \cite{mahajan2018exploring}, pre-trained in weakly-supervised fashion on 940 million public images with 1.5K hashtags matching with 1000 ImageNet1K synsets.
It is accurate (85.4\% Top-1) with lot of parameters (829 M).

\paragraph{Training protocol.}

We train ResNet-50 with SGD with a learning rate of $0.1 \times B / 256$, where $\bs$ is the batch size, as in~\cite{Tong2018BagofTricks}.
The learning rate is divided by 10 every 30 epochs.
With a Repeated Augmentation of 3, an epoch processes $5005 \times 512 / \bs$ batches, or $\sim$90\% of the training images, see~\cite{berman2019multigrain}.
In the initial training, we use $B=512$, 120 epochs and the default PyTorch data augmentation: horizontal flip, random resized crop (as in~\cref{sec:analysis}) and color jittering.
To finetune, the initial learning rate is $0.008$ same decay, $B=512$, 60 epochs. 
The data-augmentation used for fine-tuning is described in the next paragraph.
For ResNeXt-101 32x48d we use the pretrained version from PyTorch hub repository~\cite{pretrainedpytorchWSL}. 
We use almost the same fine-tuning as for the ResNet-50.
We also use a ten times smaller learning rate and a batch size two times smaller.
For PNASNet-5-Large we use the pretrained version from Cadene's GitHub repository~\cite{pretrainedpytorch}. 
The difference with the ResNet-50 fine-tuning is that we modify the last three cells, in one epoch and with a learning rate of 0.0008.
We run our experiments on machines with 8 Tesla V100 GPUs and 80\,CPU cores to train and fine-tune our ResNet-50.
\paragraph{Fine-tuning data-augmentation.}
\label{par:finetunedata}
We experimented three data-augmentation for fine-tuning:
The first one (test DA) is resizing the image and then take the center crop,
The second one (test DA2) is resizing the image, random horizontal shift of the center crop, horizontal flip and color jittering.
The last one (train DA) is the train-time data-augmentation as described in the previous paragraph.

A comparison of the performance of these data augmentation is made in the section~\ref{sec:resulttables}.

The test DA data-augmentation described in this paragraph being the simplest.
Therefore test DA is used for all the results reported with ResNet-50 and PNASNet-5-Large in this paper except in Table~\ref{tab:sota} where we use test DA2 to have slightly better performances in order to compare ours results with the state of the art.

For ResNeXt-101 32x48d all reported results are obtained with test DA2.
We make a comparison of the results obtained between testDA, testDA2 and train DA in section~\ref{sec:resulttables}.

\begin{figure*}[t]
\includegraphics[width=0.45\linewidth]{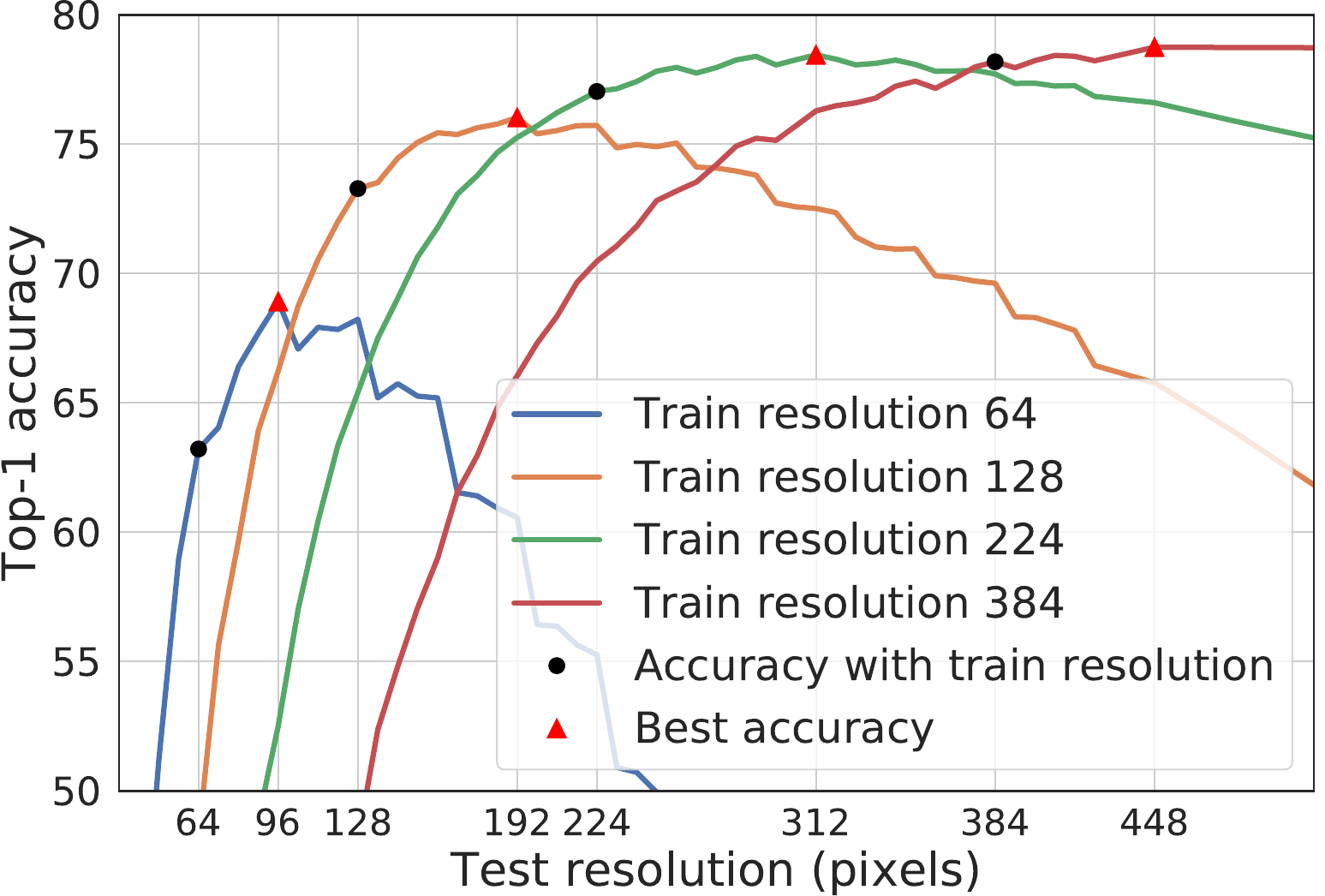}~%
\hfill
\includegraphics[width=0.45\linewidth]{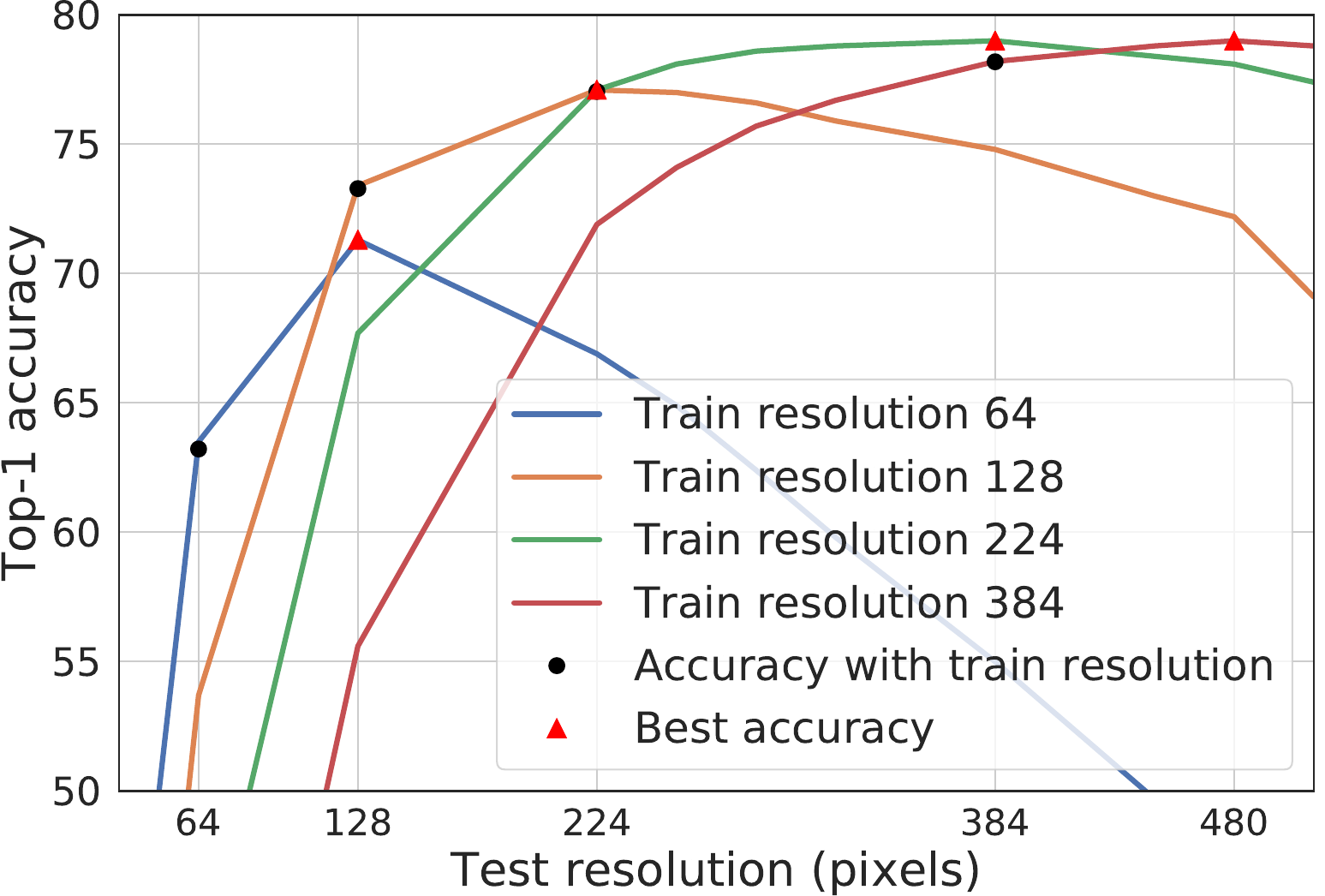}
\caption{\label{fig:accuracyresolution}
    Top-1 accuracy of the ResNet-50 according to the test time resolution.
    Left: without adaptation, right: after resolution adaptation. 
    The numerical results are reported in Appendix~\ref{sec:resulttables}.
    A comparison of results without random resized crop is reported in Appendix~\ref{sec:RandomResizedCropEffect}.
}
\end{figure*}

\paragraph{The baseline} experiment is to increase the resolution without adaptation.
Repeated augmentations already improve the default PyTorch ResNet-50 from 76.2\% top-1 accuracy to 77.0\%.
Figure~\ref{fig:accuracyresolution}(left) shows that increasing the resolution at test time increases the accuracy of all our networks. E.g., the accuracy of a ResNet-50 trained at resolution 224 increases from 77.0 to 78.4 top-1 accuracy, an improvement of 1.4 percentage points. This concurs with prior findings in the literature \cite{He2016IdentityMappings}.

\subsection{Results}

\paragraph{Improvement of our approach on a ResNet-50.} 
Figure~\ref{fig:accuracyresolution}(right) shows the results obtained after fine-tuning the last batch norm in addition to the classifier. 
With fine-tuning we get the best results ($79\%$) with the classic ResNet-50 trained at $\HWptr=224$.
Compared to when there is no fine-tuning, the $\HWpte$ at which the maximal accuracy is obtained increases from $\HWpte=288$ to $384$.
If we prefer to reduce the training resolution, $\HWptr=128$ and testing at $\HWptr=224$ yields 77.1\% accuracy, which is above the baseline trained at full test resolution without fine-tuning. 

\paragraph{Multiple resolutions.}

To improve the accuracy, we classify the image at several resolutions and average the classification scores. 
Thus, the training time remains the same but there is a modest increase in inference time compared to processing only the highest-resolution crop. 
With $\HWptr=128$ and $\HWpte=[256,192]$, the accuracy is 78.0\%.
With $\HWptr=224$ and $\HWpte=[384,352]$, we improve the single-crop result of 79.0\% to 79.5\%. 

\paragraph{Application to  larger networks.}

The same adaptation method can be applied to any convolutional network. 
In Table~\ref{tab:largenetwork} we report the result on the PNASNet-5-Large and the IG-940M-1.5k ResNeXt-101 32x48d \cite{mahajan2018exploring}. 
For the PNASNet-5-Large, we found it beneficial to fine-tune more than just the batch-normalization and the classifier. 
Therefore, we also experiment with fine-tuning the three last cells. 
By increasing the resolution to $\HWpte=480$, the accuracy increases by 1 percentage point. 
By combining this with an ensemble of 10 crops at test time, we obtain \textbf{83.9\%} accuracy. 
With the ResNeXt-101 32x48d  increasing the resolution to $\HWpte=320$, the accuracy increases by 1.0 percentage point. 
We thus reached \textbf{86.4\%} top-1 accuracy.
\def \mysp {\hspace{11pt}}

\begin{table*}
\caption{\label{tab:largenetwork}
    Application to larger networks: 
    Resulting top-1 accuracy %
  }  
  \centering {\small
  \begin{tabular}{cccccc@{\mysp}c@{\mysp}c@{\mysp}c@{\mysp}c@{\mysp}c}
    \toprule
    
    \multicolumn{1}{c}{Model} &\multicolumn{1}{c}{Train} & \multicolumn{3}{c}{Fine-tuning}& \multicolumn{6}{c}{Test resolution} \\
    \cmidrule(lr){3-5} \cmidrule(lr){6-11} 
    used &resolution & Classifier & Batch-norm & Three last Cells & 331 & 384 & 395 & 416 & 448 & 480\\
    \cmidrule(lr){1-1} \cmidrule(lr){2-2} \cmidrule(lr){3-5} \cmidrule(lr){6-11} 
    PNASNet-5-Large & 331 & \_    & \_ &  \_     &  82.7  & 83.0 & \textbf{83.2} &83.0 & 83.0 & 82.8 \\

    PNASNet-5-Large & 331 & \checkmark & \checkmark  &  \_  &  82.7 & 83.4 & 83.5 & 83.4 &\textbf{ 83.5} & 83.4 \\
    PNASNet-5-Large & 331 & \checkmark & \checkmark  &  \checkmark  &  82.7 & 83.3 & 83.4 & 83.5 & 83.6& \textbf{83.7} \\
    \midrule	
      &  & Classifier & Batch-norm & Three last conv layer & \multicolumn{2}{c}{224} & \multicolumn{2}{c}{288} & \multicolumn{2}{c}{320}\\
        \cmidrule(lr){3-5} \cmidrule(lr){6-11} 
        ResNeXt-101 32x48d   & 224 & \checkmark & \checkmark  &  \_  &  \multicolumn{2}{c}{85.4} & \multicolumn{2}{c}{86.1} & \multicolumn{2}{c}{\textbf{86.4}}\\
    \bottomrule
\end{tabular}}
\end{table*}

\paragraph{Speed-accuracy trade-off.}
\label{sec:timings}

We consider the trade-off between training time and accuracy (normalized as if it was run on 1~GPU). 
The full table with timings are in supplementary Section~\ref{sec:supptimings}.
In the initial training stage, the forward pass is 3 to 6 times faster than the backward pass. 
However, during fine-tuning the ratio is inverted because the backward pass is applied only to the last layers.

In the low-resolution training regime ($\HWptr=128$), the additional fine-tuning required by our method increases the training time from 111.8~h to 124.1~h (+11\%). 
This is to obtain an accuracy of 77.1\%, which outperforms the network trained at the native resolution of 224 in 133.9~h.
We produce a fine-tuned network with $\HWpte=384$ that obtains a higher accuracy than the network trained natively at that resolution, and the training is $2.3\times$ faster: 151.5~h instead of 348.5~h.

\begin{table*}
\caption{\label{tab:sota}
  State of the art on ImageNet with ResNet-50 architectures and with all types of architecture (Single Crop evaluation)
}
\centering
\small
\begin{tabular}{lcccccc}
  \toprule
Models & Extra Training Data &Train  & Test   & \# Parameters & Top-1 (\%) &  Top-5 (\%) \\
\midrule	
ResNet-50 Pytorch                                &   \_     & 224   & 224                     & 25.6M     & 76.1      & 92.9\\
ResNet-50 mix up \cite{Zhang2017Mixup}           &   \_    & 224   & 224                     & 25.6M     & 77.7      & 94.4\\
ResNet-50 CutMix  \cite{Yun2019CutMix}           &   \_    & 224   & 224                     & 25.6M     & 78.4      & 94.1\\
ResNet-50-D \cite{Tong2018BagofTricks}           &   \_     & 224   & 224                     & 25.6M     & 79.3      & 94.6\\
MultiGrain R50-AA-500 \cite{berman2019multigrain}&   \_    & 224   & 500                     & 25.6M    & 79.4      & 94.8\\
ResNet-50 Billion-scale \cite{Yalniz2019BillionscaleSL}             &   \checkmark    & 224   & 224                     & 25.6M    & 81.2      & 96.0\\
\midrule

Our ResNet-50                                   &  \_      & 224   & 384                     & 25.6M    & 79.1      & 94.6\\
Our ResNet-50 CutMix                            &  \_     & 224   & 320                     & 25.6M    & 79.8      & 94.9\\
Our ResNet-50 Billion-scale@160                &   \checkmark     & 160   & 224                     & 25.6M    & 81.9      & 96.1\\
Our ResNet-50 Billion-scale@224                &   \checkmark     & 224   & 320                     & 25.6M    & 82.5      & 96.6\\

\midrule
PNASNet-5 (N                                             = 4, F      = 216) \cite{Liu2018PNAS} & \_   & 331   & 331   & 86.1M & 82.9 & 96.2\\
  MultiGrain PNASNet @ 500px \cite{berman2019multigrain} & & 331   & 500                     & 86.1M     & 83.6      & 96.7\\
AmoebaNet-B (6,512) \cite{Yanping2018GPipe}      &  \_      & 480   & 480                     & 577M     & 84.3      & 97.0\\
EfficientNet-B7  \cite{tan2019efficientnet}      &   \_     & 600   & 600                     & 66M     & 84.4      & 97.1\\

\midrule
  Our PNASNet-5                                  &  \_     & 331   & 480                     & 86.1M     & 83.7      & 96.8 \\
\midrule
    ResNeXt-101 32x8d \cite{mahajan2018exploring}                                  &  \checkmark     & 224   & 224                     & 88M     & 82.2      & 96.4 \\
    ResNeXt-101 32x16d \cite{mahajan2018exploring}                                 &  \checkmark     & 224   & 224                     & 193M     & 84.2      & 97.2 \\
    ResNeXt-101 32x32d \cite{mahajan2018exploring}                                 &  \checkmark    & 224   & 224                     & 466M     & 85.1      & 97.5 \\
    ResNeXt-101 32x48d  \cite{mahajan2018exploring}                                &  \checkmark     & 224   & 224                     & 829M     & 85.4      & 97.6 \\
 \midrule
    Our ResNeXt-101 32x48d                             &  \checkmark     & 224   & 320                     & 829M     & \textbf{86.4}      &  \textbf{98.0} \\

  \bottomrule
\end{tabular}
\vspace{-1em}
\end{table*}

\paragraph{Ablation study.}\label{sec:ablation}

We study the contribution of the different choices to the performance, limited to $\HWptr=128$ and $\HWptr=224$.
By simply fine-tuning the classifier (the fully connected layers of ResNet-50) with test-time augmentation, we reach 78.9\% in Top-1 accuracy with the classic ResNet-50 initially trained at resolution 224. 
The batch-norm fine-tuning and improvement in data augmentation advances it to 79.0\%.
The higher the difference in resolution between training and testing, the more important is batch-norm fine-tuning to adapt to the data augmentation. 
The full results are in the supplementary Section~\ref{sec:resulttables}.

\subsection{Beyond the current state of the art}

Table~\ref{tab:sota} compares our results with competitive methods from the literature. 
Our ResNet-50 is slightly worse than ResNet50-D and MultiGrain, but these do not have exactly the same architecture.
On the other hand our ResNet-50 CutMix, which has a classic ResNet-50 architecture, outperforms others ResNet-50 including the slightly modified versions.
Our fine-tuned PNASNet-5 outperforms the MultiGrain version.
To the best of our knowledge our ResNeXt-101 32x48d  surpasses all other models available in the literature.

With \textbf{86.4\%} Top-1 accuracy and \textbf{98.0\%} Top-5 accuracy it is the first model to exceed 86.0\% in Top-1 accuracy and  98.0\% in Top-5 accuracy on the ImageNet-2012 benchmark~\cite{Russakovsky2015ImageNet12}.
It exceeds the previous state of the art~\cite{mahajan2018exploring} by 1.0\% absolute in Top-1 accuracy and 0.4\% Top-5 accuracy.

\subsection{Transfer learning tasks}
\label{sec:transferlearning}

We have used our method in transfer learning tasks to validate its effectiveness on other dataset than ImageNet. We evaluated it on the following datasets:
iNaturalist 2017~\cite{Horn2017INaturalist}, Stanford Cars~\cite{Cars2013}, CUB-200-2011~\cite{WahCUB_200_2011}, Oxford 102 Flowers~\cite{Nilsback08}, Oxford-IIIT Pets~\cite{parkhi12a}, NABirds~\cite{Horn2015NaBirds} and Birdsnap~\cite{Berg2014Birdsnap}.
We used our method with two types of networks for transfer learning tasks: SENet-154~\cite{Hu2017SENet} and InceptionResNet-V2~\cite{Szegedy2016InceptionResNetV2}.

For all these experiments, we proceed as follows.
(1) we initialize our network with the weights learned on ImageNet (using models from \cite{pretrainedpytorch}).
(2) we train it entirely for several epochs at a certain resolution. 
(3) we fine-tune with a higher resolution the last batch norm and the fully connected layer. 

Table~\ref{tab:transfertask} summarizes the models we used and the performance we achieve.
We can see that in all cases our method improves the performance of our baseline.
Moreover, we notice that the higher the image resolution, the more efficient the method is.
This is all the more relevant today, as the quality of the images increases from year to year. 

\begin{table*}
\caption{\label{tab:transfertask}
 Transfer learning task with our method and comparison with the state of the art.
We only compare ImageNet-based transfer learning results with a single center crop for the evaluation (if available, otherwise we report the best published result) without any change in architecture compared to the one used on ImageNet.
We report Top-1 Accuracy(\%).
}
\centering
\small
\begin{tabular}{llcc|lc}
  \toprule
    Dataset & Models  & Baseline   & With Our Method &  \multicolumn{2}{c}{State-Of-The-Art Models} \\
\midrule
	iNaturalist 2017~\cite{Horn2017INaturalist} & SENet-154         & 74.1 & \textbf{75.4}  & IncResNet-V2-SE \cite{Horn2017INaturalist} & 67.3 \\
	Stanford Cars~\cite{Cars2013}               & SENet-154         & 94.0 & 94.4 & EfficientNet-B7  \cite{tan2019efficientnet}  & \textbf{94.7} \\
	CUB-200-2011~\cite{WahCUB_200_2011}         & SENet-154         & 88.4 & \textbf{88.7} & 	MPN-COV \cite{Li2017MPN} & \textbf{88.7}\\
	Oxford 102 Flowers~\cite{Nilsback08}        & InceptionResNet-V2& 95.0 & 95.7 &  EfficientNet-B7  \cite{tan2019efficientnet}  & \textbf{98.8}\\
	Oxford-IIIT Pets~\cite{parkhi12a}           & SENet-154         & 94.6 & 94.8 & AmoebaNet-B (6,512) \cite{Yanping2018GPipe}  &\textbf{95.9}\\
	NABirds~\cite{Horn2015NaBirds}              & SENet-154         & 88.3 & \textbf{89.2} & PC-DenseNet-161 \cite{Dubey2017PCDenseNet} & 82.8 \\
	Birdsnap~\cite{Berg2014Birdsnap}            & SENet-154         & 83.4 & \textbf{84.3}&  EfficientNet-B7  \cite{tan2019efficientnet}  & \textbf{84.3}\\
  \bottomrule
\end{tabular}
\end{table*}

\section{Conclusion}

We have studied extensively the effect of using different train and test scale augmentations on the statistics of natural images and of the network's pooling activations.
We have shown that, by adjusting the crop resolution and via a simple and light-weight parameter adaptation, it is possible to increase the  accuracy of standard classifiers significantly, everything being equal otherwise.
We have also shown that researchers waste resources when both training and testing strong networks at resolution $224\times 224$;
We introduce a method that can ``fix'' these networks  post-facto and thus improve their performance.
An open-source implementation of our method is available at \url{https://github.com/facebookresearch/FixRes}.

\bibliographystyle{plain}
\bibliography{egbib,more}
\clearpage \newpage %
\appendix%

\onecolumn
\begin{minipage}{\textwidth}
\begin{center}
\Huge
Supplementary material for \\
``Fixing the train-test resolution discrepancy''
\end{center}
\end{minipage}

\vspace*{1cm}

In this supplementary material we report details and results that did not fit in the main paper.
This includes the estimation of the parametric distribution of activations in Section~\ref{sec:parametricfit}, a small study on border/round-off effects of the image size for a convolutional neural net in Section~\ref{sec:roundoff} and more exhaustive result tables in Section~\ref{sec:resulttables}. Section~\ref{sec:challenges} further demonstrates the interest of our approach through our participation to two competitive challenges in fine-grained recognition.

\section{Fitting the activations}
\label{sec:parametricfit}

\subsection{Parametric Fr\'echet model after average-pooling}
In this section we derive a parametric model that fits the distribution of activations on output of the spatial pooling layer. 

The output the the last convolutional layer can be well approximated with a Gaussian distribution. 
Then the batch-norm centers the Gaussian and reduces its variance to unit, and the ReLU replaces the negative part with 0. 
Thus the ReLU outputs an equal mixture of a cropped unit Gaussian and a Dirac of value 0.

The average pooling sums $n=2\times2$ to $n=14\times14$ of those distributions together.
Assuming independence of the inputs, it can be seen as a sum of $n'$ cropped Gaussians, where $n'$ follows a discrete binomial distribution. 
Unfortunately, we found this composition of distributions is not tractable in close form. 

Instead, we observed experimentally that the output distribution is close to an extreme value distribution. 
This is due to the fact that only the positive part of the Gaussians contributes to the output values.
In an extreme value distribution that is the sum of several (arbitrary independent) distributions, the same happens: only the highest parts of those distributions contribute. 

Thus, we model the statistics of activations as a Fr\'echet (a.k.a. inverse Weibull) distribution. 
This is a 2-parameter distribution whose CDF has the form: 
\[
P(x,\mu,\sigma) = e^{-\left(1+\frac{\xi}{\sigma}(x-\mu)\right)^{-1/\xi}}
\]
With~$\xi$~a~positive~constant, $\mu \in \mathbb{R} , \sigma \in \mathbb{R}^{*}_{+}.$ 
We observed that the parameter $\xi$ can be kept constant at $0.3$ to fit the distributions.

\begin{figure*}[b]
\begin{minipage}{0.24\linewidth}
    \includegraphics[width=\linewidth]{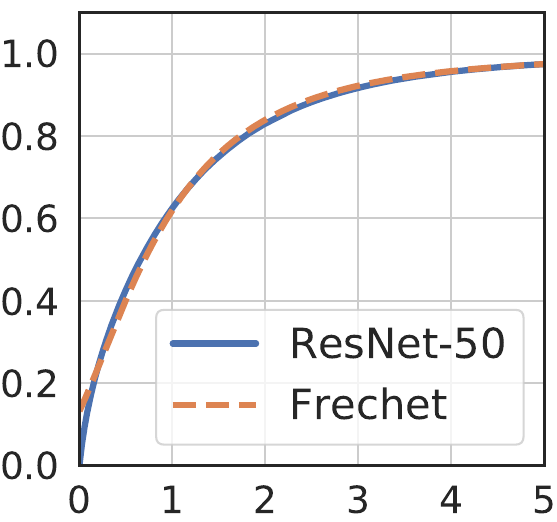}  \\
    Resolution: 64
\end{minipage}
\hfill
\begin{minipage}{0.24\linewidth}
\includegraphics[width=\linewidth]{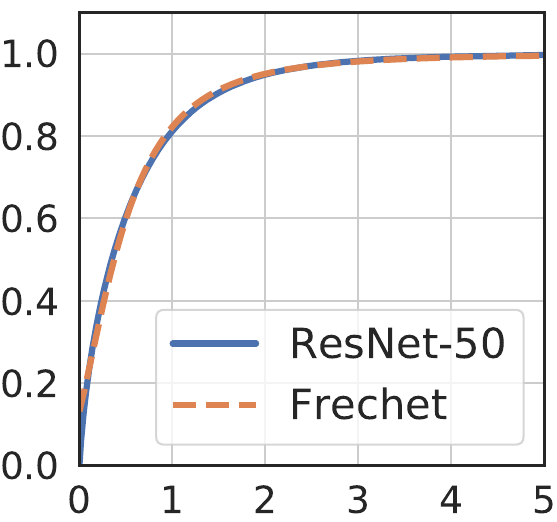} \\
Resolution: 128
\end{minipage}
\hfill
\begin{minipage}{0.24\linewidth}
\includegraphics[width=\linewidth]{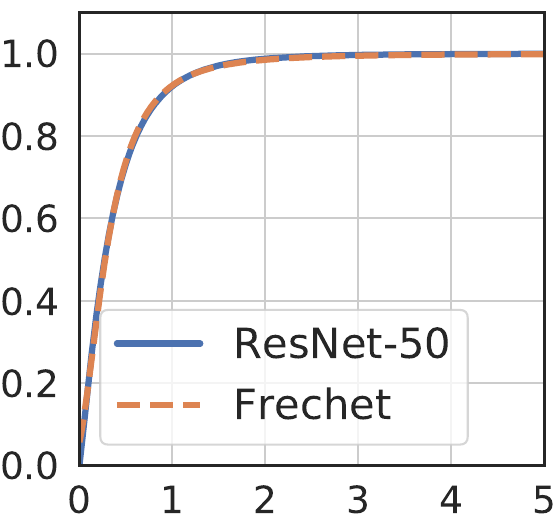} \\
Resolution: 224
\end{minipage}
\hfill
\begin{minipage}{0.24\linewidth}
\includegraphics[width=\linewidth]{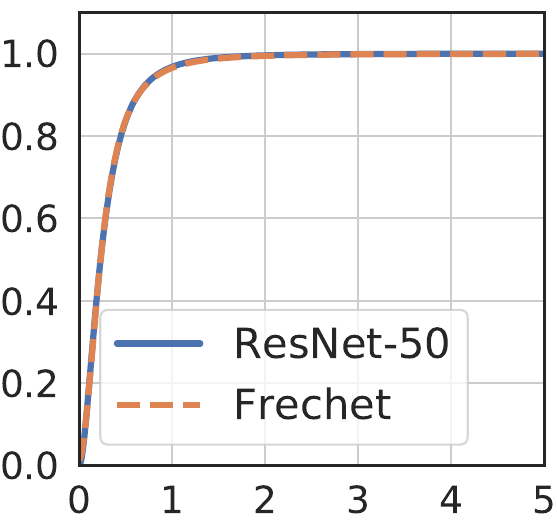} \\
 Resolution: 448
 \end{minipage}   
\caption{\label{fig:pdfresolutionfrechet}
    Fitting of the CDF of activations with a Fr\'echet distribution.
}
\end{figure*}

Figure~\ref{fig:pdfresolutionfrechet} shows how the Fr\'echet model fits the empirical CDF of the distribution. 
The parameters were estimated using least-squares minimization, excluding the zeros, that can be considered outliers.
The fit is so exact that the difference between the curves is barely visible.

To correct the discrepancy in distributions at training and test times, we compute the parameters $\mu_\mathrm{ref}, \sigma_\mathrm{ref}$ of the distribution observed on training images time for $\HWpte = \HWptr$. 
Then we increase $\HWpte$ to the target resolution and measure the parameters $\mu_0, \sigma_0$ again. 
Thus, the transformation is just an affine scaling, still ignoring zeros.

When running the transformed neural net on the Imagenet evaluation, we obtain accuracies: 
\begin{center}
\begin{tabular}{|l|rrrrrr|}
\hline 
$\HWcte$ & 64 & 128 & 224 & 256 & 288 & 448 \\
    accuracy & 29.4 & 65.4 & 77 &78 & 78.4 & 76.5 \\
\hline
\end{tabular}
\end{center}
Hence, the accuracy does not improve with respect to the baseline. 
This can be explained by several factors: 
the scalar distribution model, however good it fits to the observations, is insufficient to account for the individual distributions of the activation values; 
just fitting the distribution may not be enough to account for the changes in behavior of the convolutional trunk.

\subsection{Gaussian model before the last ReLU activation}

Following the same idea as what we did previously we looked at the distribution of activations by channel  before the last ReLU according to the resolution.

We have seen that the distributions are different from one resolution to another. 
With higher resolutions,  the mean tends to be closer to 0 and the variance tends to become smaller.
By acting on the distributions before the ReLU, it is also possible to affect the sparsity of values after spatial-pooling, which was not possible with the previous analysis based on Frechet's law.
We aim at matching the distribution before the last ReLU with the distribution of training data at lower resolution. 
We compare the effect of this transformation before/after fine tuning with the learnt batch-norm approach. 
The results are summarized in Table~\ref{tab:adaptlaw}.

We can see that adapting the resolution by changing the distributions is effective especially in the case of small resolutions. 
Nevertheless, the adaptation obtained by fine-tuning the the batch norm improves performs better in general.

\begin{table*}
\caption{\label{tab:adaptlaw}
    Matching  distribution before the last Relu  application to  ResNet-50: 
    Resulting top-1 accuracy \% on ImageNet validation set
  }  
  \centering {\small
  \begin{tabular}{cccccc@{\mysp}c@{\mysp}c@{\mysp}c@{\mysp}c}
    \toprule
    
    \multicolumn{1}{c}{Model} &\multicolumn{1}{c}{Train}  & \multicolumn{1}{c}{Adapted }& \multicolumn{2}{c}{Fine-tuning}& \multicolumn{5}{c}{Test resolution} \\
     \cmidrule(lr){4-5} \cmidrule(lr){6-10} 
    used &resolution  & Distribution & Classifier & Batch-norm & 64 & 224 & 288 & 352 & 384\\
    \cmidrule(lr){1-1} \cmidrule(lr){2-2} \cmidrule(lr){3-5} \cmidrule(lr){6-10} 
    ResNet-50 & 224 & \_    & \_ &  \_     &  29.4  & 77.0 & 78.4 & 78.1 & 77.7  \\
    ResNet-50 & 224 & \checkmark    & \_ &  \_     &  29.8  & 77.0 & 77.7 & 77.3 & 76.8 \\
    ResNet-50 & 224 & \_    & \checkmark  &  \_     &  40.6  & 77.1 & 78.6 &78.9 & 78.9 \\
    ResNet-50 & 224 & \_    & \checkmark  &  \checkmark   &  41.7  & 77.1 & 78.5 & 78.9 & 79.0  \\
    ResNet-50 & 224 & \checkmark    & \checkmark  &  \_   &  41.8  & 77.1 & 78.5 & 78.8 & 78.9 \\

    \bottomrule
\end{tabular}}
\end{table*}

\section{Border and round-off effects}
\label{sec:roundoff}

Due to the complex discrete nature of convolutional layers, the accuracy is not a monotonous function of the input resolution. 
There is a strong dependency on the kernel sizes and strides used in the first convolutional layers. 
Some resolutions will not match with these parameters so we will have a part of the images margin that will not be taken into account by the convolutional layers. 

\begin{figure}[t]
\begin{center}
\includegraphics[width=0.5\columnwidth]{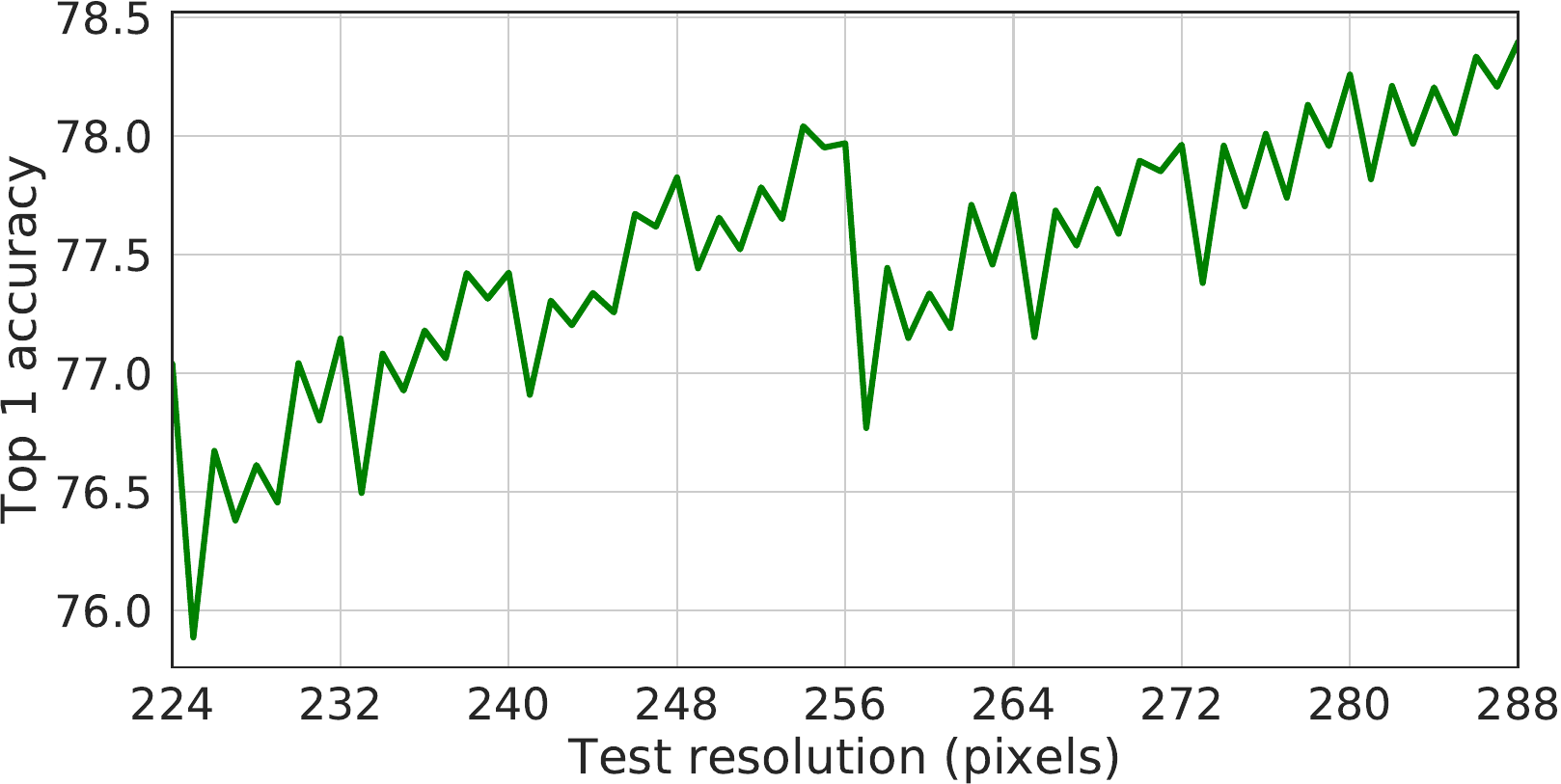}~%
\end{center}
\caption{\label{fig:onepixels}
    Evolution of the top-1 accuracy of the ResNet-50 trained with resolution 224 according to the testing resolution (no finetuning).
    This can be considered a zoom of figure~\ref{fig:accuracyresolution} with 1-pixel increments.
}
\end{figure}

In Figure~\ref{fig:onepixels}, we show the variation in accuracy when the resolution of the crop is increased by steps of 1~pixel.  
Of course, it is possible to do padding but it will never be equivalent to having a resolution image adapted to the kernel and stride size.

Although the global trend is increasing, there is a lot of jitter that comes from those border effects. 
There is a large drop just after resolution 256.
We observe the drops at each multiple of 32, they correspond to a changes in the top-level activation map's resolution.
Therefore we decided to use only sizes that are multiples of 32 in the experiments. 

\section{Result tables}
\label{sec:resulttables}

Due to the lack of space, we report only the most important results in the main paper. 
In this section, we report the full result tables for several experiments.

Table~\ref{tab:baselines} report the numerical results corresponding to Figure~\ref{fig:accuracyresolution} in the main text.
Table~\ref{tab:ablationstudytable} reports the full ablation study results (see Section~\ref{sec:ablation}). 
\label{sec:supptimings}
Table~\ref{tab:time} reports the runtime measurements that Section~\ref{sec:timings} refers to.
Table~\ref{tab:dataaugmentationcomparison} reports a comparaison between test DA and test DA2 that Section~\ref{sec:experiments} refers to. 

\begin{table*}

{\small ~\hfill \begin{tabular}{lrcccc}
    \toprule
    test $\backslash$ train     & 64 \ \ & 128 & 160 & 224 & 384 \\
    \midrule
    64 & 63.2 & 48.3 & 40.1 & 29.4 & 12.6 \\
    128 & \textbf{68.2} & 73.3 &71.2 & 65.4 & 48.0 \\
    224 & 55.3 & \textbf{75.7} &\textbf{ 77.3}  & 77.0 & 70.5 \\
    288 & 42.4 & 73.8 & 76.6 & \textbf{78.4} & 75.2 \\
    384 & 23.8 & 69.6 & 73.8 & 77.7 & 78.2 \\
    448 & 13.0 & 65.8 &71.5 & 76.6 & \textbf{78.8} \\
    480 & 9.7 & 63.9 & 70.2 & 75.9 & 78.7 \\
    \bottomrule
  \end{tabular}%
  \hfill
  \begin{tabular}{l|llll}
    \toprule
    test $\backslash$ train     & 64 & 128 & 224 & 384 \\
    \midrule
    64 & 63.5  & 53.7  & 41.7 & 27.5 \\
    128 & \textbf{71.3} & 73.4 & 67.7 & 55.7 \\
    224 & 66.9 & \textbf{77.1}  & 77.1 & 71.9 \\
    288 & 62.4 & 76.6 & 78.6 & 75.7 \\
    384 & 55.0 & 74.8 & \textbf{79.0} & 78.2 \\
    448 & 49.7 & 73.0 & 78.4 & 78.8 \\
    480 & 46.6 & 72.2 & 78.1 & \textbf{79.0} \\
    \bottomrule
\end{tabular} \hfill ~}
\smallskip
    \caption{
    \label{tab:baselines}
	Top-1 validation accuracy for different combinations of training and testing resolution. 
	Left: with the standard training procedure, (no finetuning, no adaptation of the ResNet-50).
	Right: with our data-driven adaptation strategy and test-time augmentations.
}
\end{table*}

\begin{table*}
\centering
  {\small
  \begin{tabular}{c|ccc|cccccc}
    \toprule
    \multicolumn{1}{c}{Train} & \multicolumn{3}{|c}{Fine-tuning}& \multicolumn{6}{|c}{Test resolution (top-1 accuracy)} \\
    \cmidrule(lr){2-4} \cmidrule(lr){5-10} 
    resolution & Classifier & Batch-norm &  Data aug. & 64 & 128 &224 & 288 & 384 & 448\\
    \midrule
        & --         & -- &  n/a     &  48.3  & 73.3 & \textbf{75.7} & 73.8 & 69.6 & 65.8 \\
        & \checkmark & -- &  train DA &  52.8 & 73.3 & \textbf{77.1} & 76.3& 73.2 & 71.7 \\
    128 & \checkmark & -- &  test DA&  53.3 & 73.4 & \textbf{77.1} & 76.4& 74.4 & 72.3 \\
        & \checkmark & \checkmark &  train DA &  53.0 & 73.3 & \textbf{77.1} & 76.5 & 74.4 & 71.9 \\
        & \checkmark & \checkmark  &  test DA &  53.7 & 73.4 & \textbf{77.1} & 76.6 & 74.8 & 73.0 \\
    \midrule
         & --         & -- &  n/a     & 29.4 & 65.4 & 77.0 & \textbf{78.4} & 77.7 & 76.6 \\
        & \checkmark & -- &  train DA &  39.9 &67.5 & 77.0 & 78.6 & \textbf{78.9} & 78.0 \\
    224 & \checkmark & -- &   test DA &  40.6 & 67.3 & 77.1 & 78.6 & \textbf{78.9} & 77.9 \\
        & \checkmark & \checkmark &  train DA  &  40.4  & 67.5 & 77.0 & 78.6 & \textbf{78.9} & 78.0 \\
        & \checkmark & \checkmark &   test DA &  41.7 & 67.7 & 77.1 & 78.6 & \textbf{79.0} & 78.4 \\
    \bottomrule
\end{tabular}}
\smallskip
\caption{\label{tab:ablationstudytable}
    Ablation study: 
    Accuracy when enabling or disabling some components of the training method. 
    Train DA: training-time data augmentation during fine-tuning, test DA: test-time one. 
}
\end{table*}

\newcommand{\std}[1]{\footnotesize{$\pm #1$}}

\begin{table*}
\centering
\small
\begin{tabular}{ll|rrrr|rc}
  \toprule    
  \multicolumn{2}{c}{Resolution} & \multicolumn{2}{c}{Train time per batch (ms)} & \multicolumn{2}{c}{Resolution fine-tuning (ms)} & \multicolumn{2}{c}{Performance} \\
  \cmidrule(lr){1-2} \cmidrule(lr){3-4} \cmidrule(lr){5-6} \cmidrule(lr){7-8} 
  train & test & backward        & forward        & backward      & forward        & Total time (h)       & accuracy  \\
  \midrule
  128 & 128 & 29.0 \std{4.0}  & 12.8 \std{2.8} & \_            & \_             & 111.8 \hspace{1.5em} & 73.3 \\
  160 & 160 & 30.2 \std{3.2}  & 14.5 \std{3.4} & \_            & \_             & 119.7 \hspace{1.5em} & 75.1 \\
  224 & 224 & 35.0 \std{2.0}  & 15.2 \std{3.2} & \_            & \_             & 133.9 \hspace{1.5em} & 77.0 \\
  384 & 384 & 112.4 \std{6.2} & 18.2 \std{3.9} & \_            & \_             & 348.5 \hspace{1.5em} & 78.2 \\
  \midrule
  160 & 224 & 30.2 \std{3.2}  & 14.5 \std{3.4} & \_            & \_             & 119.7 \hspace{1.5em} & 77.3 \\
  224 & 288 & 35.0 \std{2.0}  & 15.2 \std{3.2} & \_            & \_             & 133.9 \hspace{1.5em} & 78.4 \\
    \midrule
  128 & 224 & 29.0 \std{4.0}  & 12.8 \std{2.8} & 4.4 \std{0.9} & 14.4 \std{2.5} & 124.1 \hspace{1.5em} & 77.1 \\
  160 & 224 & 30.2 \std{3.2}  & 14.5 \std{3.4} & 4.4 \std{0.9} & 14.4 \std{2.5} & 131.9 \hspace{1.5em} & 77.6  \\
  224 & 384 & 35.0 \std{2.0}  & 15.2 \std{3.2} & 8.2 \std{1.3} & 18.0 \std{2.7} & 151.5 \hspace{1.5em} & 79.0\\
  \bottomrule
\end{tabular}
\smallskip
\caption{\label{tab:time}
Execution time for the training. 
Training and fine-tuning times are reported for a batch of size 32 for training and 64 for fine-tuning, on one GPU.  
Fine-tuning uses less memory than training therefore we can use larger batch size.
The total time is the total time spent on both, with 120 epochs for training and 60 epochs of fine-tuning on ImageNet. 
Our approach corresponds to fine-tuning of the batch-norm and the classification layer.
}
\end{table*}

\begin{table*}
\centering
\small
\begin{tabular}{lcccc}
    \toprule 
    Models & Train  & Test & Top-1 test DA (\%) & Top-1 test DA2 (\%) \\
    \midrule
    ResNext-101 32x48d  & 224 & 288 & 86.0 & 86.1 \\
    ResNext-101 32x48d & 224 & 320 & 86.3 & 86.4 \\
    \midrule
    ResNet-50  & 224 & 320 & 79.0 & 79.1 \\
    \midrule
    ResNet-50 CutMix & 224 & 384 & 79.7 & 79.8  \\
    \bottomrule
\end{tabular}
\smallskip
\caption{\label{tab:dataaugmentationcomparison}
Comparisons of performance between data-augmentation test DA and test DA2 in the case of fine-tuning batch-norm and classifier.
}
\end{table*}

\section{Impact of Random Resized Crop}
\label{sec:RandomResizedCropEffect}
\begin{figure}[t]
\begin{center}
\includegraphics[width=0.5\columnwidth]{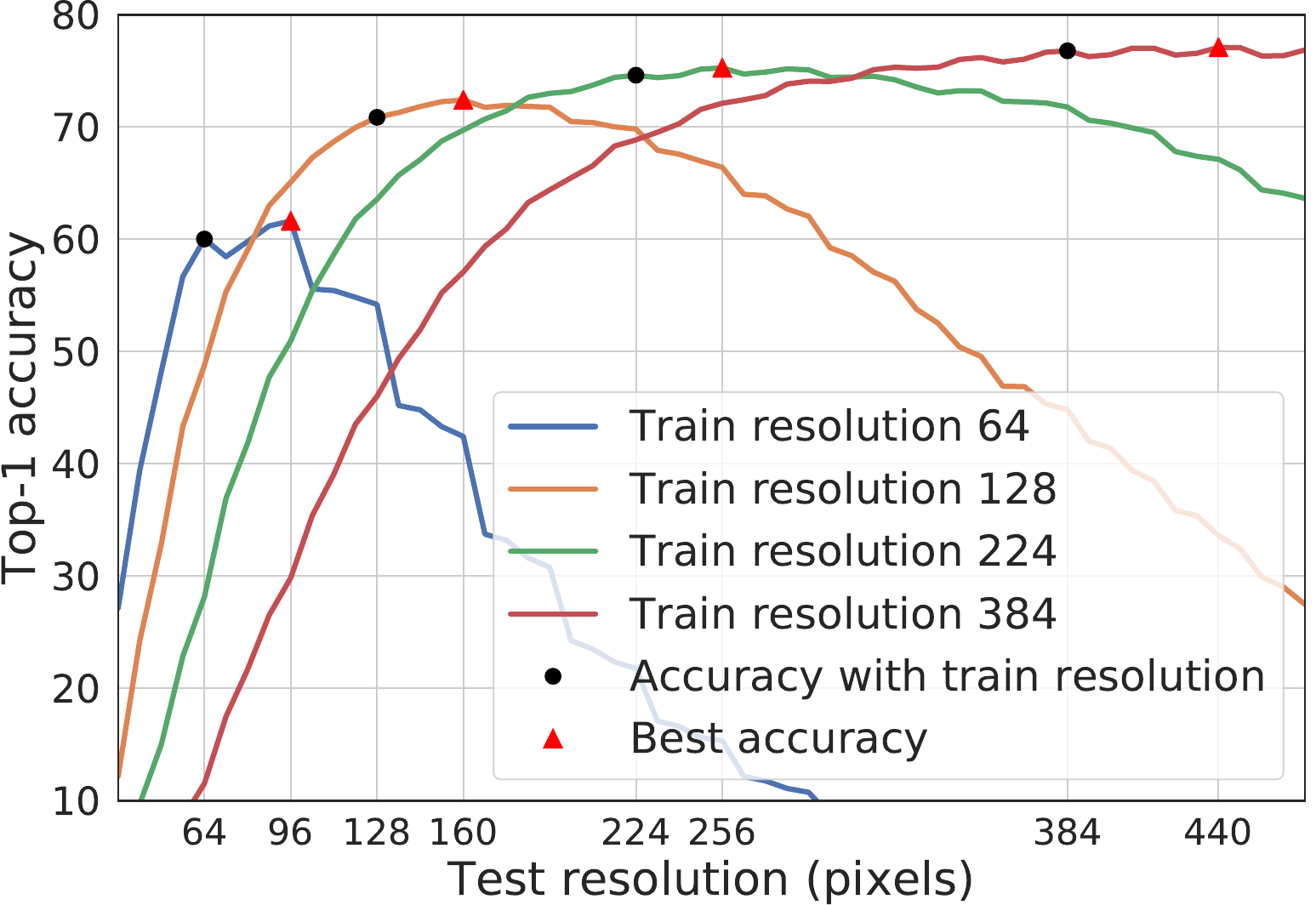}~%
\end{center}
\caption{\label{fig:norandomresizedcrop}
Top-1 accuracy of the ResNet-50 according to the test time resolution. ResNet-50 train with resize and random crop with a fixed size instead of random resized crop.
}
\end{figure}

In this section we measure the impact of the RandomResizedCrop illustrated in the section~\ref{sec:experiments}.
To do this we did the same experiment as in section~\ref{sec:experiments} but we replaced the RandomResizedCrop with a Resize followed by a random crop with a fixed size.
The figure~\ref{fig:norandomresizedcrop} and table~\ref{tab:norandomresizedcrop} shows our results. 
We can see that the effect observed in the section~\ref{sec:experiments} is mainly due to the Random Resized Crop as we suggested with our analysis of the section~\ref{sec:analysis}.

\begin{table}
  \centering
  \begin{tabular}{l|llll}
    \toprule
    test $\backslash$ train     & 64 & 128 & 224 & 384 \\
    \midrule
    64 & 60.0 & 48.7 & 28.1 & 11.5 \\
    96 & \textbf{61.6} & 65.0 & 50.9 & 29.8 \\
    128 & 54.2 & 70.8 & 63.5 & 46.0 \\
    160 & 42.4 & \textbf{72.4} & 69.7 & 57.0 \\
    224 & 21.7 & 69.8 & 74.6 & 68.8 \\
    256 & 15.3 & 66.4 & \textbf{75.2} & 72.1 \\
    384 & 4.3 & 44.8 & 71.7 & 76.7 \\
    440 & 2.3 & 33.6 & 67.1 & \textbf{77.0} \\
    \bottomrule
\end{tabular}
\caption{\label{tab:norandomresizedcrop}
	Top-1 validation accuracy for different combinations of training and testing resolution. 
ResNet-50 train with resize and random crop with a fixed size instead of random resized crop.
}
\end{table}

\section{Fine-Grained Visual Categorization contests: iNaturalist \& Herbarium}
\label{sec:challenges}

In this section we summarize the results we obtained with our method during the CVPR 2019 iNaturalist~\cite{Horn2017INaturalist} and Herbarium~\cite{Kiat2019Herbarium} competitions\footnote{
\url{https://www.kaggle.com/c/herbarium-2019-fgvc6} \newline
\url{https://www.kaggle.com/c/inaturalist-2019-fgvc6}}.
We used the approach lined out in Subsection~\ref{sec:transferlearning}, except that we adapted the preprocessing to each dataset and added a few ``tricks'' useful in competitions (ensembling, multiple crops). 

\subsection{Challenges}

The iNaturalist Challenge 2019 dataset contains images of 1010 animal and plant species, with a training set of 268,243 images and a test set of 35,351 images. 
The main difficulty is that the species are very similar within the six main families (Birds, Reptiles, Plants, Insects, Fungi and Amphibians) contained in the dataset. 
There is also a very high variability within the classes as the appearance of males, females and juveniles is often very different.
What also complicates the classification is the size of the area of interest which is very variable from one image to another, sometimes the images are close-ups on the subject, sometimes we can hardly distinguish it. 
As a preprocessing, all images have been resized to have a maximum dimension of 800 pixels.

The Herbarium contest requires to identify melastome species from 683 herbarium specimenina.
The training set contain 34,225 images and the test set contain 9,565 images. 
The main difficulty is that the specimina are very similar and not always intact. 
In this challenge the particularity is that there is no variability in the background: each specimen is photographed on a white sheet of paper. All images have been also resized to have a maximum dimension of 800 pixels.

\subsection{Ensemble of classifiers}

In both cases we used 4 different CNNs to do the classification and we averaged their results, which are themselves from 10 crops of the image. 
We chose 4 quite different in their architectures in order to obtain orthogonal classification results.
We tried to include the ResNet-50, but it was significantly worse than the other models, even when using an ensemble of models, probably due to its limited capacity. 

We used two fine-tuning stages: (1) to adapt to the new dataset in 120 epochs and (2) to adapt to a higher resolution in a few epochs. 
We chose the initial training resolution with grid-search, within the computational constraints. 
We did not skew the sampling to balance the classes. 
The rationale for this is that the performance measure is top-1 accuracy, so the penalty to misclassify infrequent classes is low.

\subsection{Results}

Table~\ref{tab:challenge} summarizes the parameters of our submission and the results. 
We report our top-performing approach, 3 and 1 points behind the winners of the competition. 
Note that we just used our method off-the-shelf and therefore used much fewer evaluations on the public part of the test set (5 for iNaturalist and 8 for Herbarium). 
The number of CNNs that at are combined in our ensemble is also smaller that two best performing ones.
In addition, for iNaturalist we did not train on data from the 2018 version of the contest.
In summary, our participation was a run with minimal if no tweaking, where we obtain excellent results (5th out of more than 200 on iNaturalist), thanks to the test-time resolution adaptation exposed in this article.

\begin{table*}
\centering
  \small
  \begin{tabular}{c|c|ccc|c}
    \toprule
    \multicolumn{1}{c}{INaturalist}& \multicolumn{1}{|c}{Train} & \multicolumn{3}{|c}{Fine-tuning} & \multicolumn{1}{|c}{Test}\\
    \cmidrule(lr){3-5} 
    Model used & resolution & Layer 4 & Classifier & Batch-norm &  resolution  \\
    \midrule
  SE-ResNext-101-32x4d & 448 &  -- & \checkmark & \checkmark &  704 \\
  SENet-154 & 448 &  \checkmark  & \checkmark & \checkmark &  672 \\
  Inception-ResNet-V2 & 491 &  -- & \checkmark & \checkmark &  681 \\
  ResNet-152-MPN-COV \cite{Li2017MPN} & 448 &  -- & -- & -- &  448 \\
   \midrule
   \multicolumn{1}{c}{}& \multicolumn{1}{c}{final score : } &\multicolumn{1}{c}{86.577 \%}& \multicolumn{1}{c}{Rank : 5 / 214} &\multicolumn{1}{c}{}& \multicolumn{1}{c}{}\\
   \toprule

    \multicolumn{1}{c}{Herbarium}& \multicolumn{1}{|c}{Train} & \multicolumn{3}{|c}{Fine-tuning} & \multicolumn{1}{|c}{Test}\\
    \cmidrule(lr){3-5} 
    Model used & resolution & Layer 4 & Classifier & Batch-norm &  resolution  \\
    \midrule
    SENet-154 & 448 &  --  & \checkmark & \checkmark &  707 \\
    ResNet-50 & 384 &  --  & \checkmark & \checkmark &  640 \\
    \midrule
   \multicolumn{1}{c}{}& \multicolumn{1}{c}{final score : } &\multicolumn{1}{c}{88.845 \%}& \multicolumn{1}{c}{Rank : 4 / 22} &\multicolumn{1}{c}{}& \multicolumn{1}{c}{}\\
    \bottomrule
\end{tabular}
\smallskip
\caption{\label{tab:challenge}
Our best ensemble results for the Herbarium and INaturalist competitions.
}
\end{table*}

\end{document}